\documentclass{article}

     \usepackage[final, nonatbib]{neurips_2020}

\usepackage[utf8]{inputenc} % allow utf-8 input
\usepackage[T1]{fontenc}    % use 8-bit T1 fonts
\usepackage{hyperref}       % hyperlinks
\usepackage{url}            % simple URL typesetting
\usepackage{booktabs}       % professional-quality tables
\usepackage{amsfonts}       % blackboard math symbols
\usepackage{nicefrac}       % compact symbols for 1/2, etc.
\usepackage{microtype}      % microtypography
\usepackage{xcolor}
\usepackage{graphics}
\usepackage{graphicx}

\usepackage{caption}
\usepackage{subcaption}

\newif\ifcomments

\commentstrue

\ifcomments
\newcommand{\comments}[1]{#1}
\else
\newcommand{\comments}[1]{}
\fi
\title{Deconstructing the Structure of Sparse Neural Networks}

\author{%
      Maxwell Van Gelder\thanks{Correspondence to \texttt{maxdvg@uw.edu}. $^\dagger$Also affiliated with the Allen Institute for AI.} \\
    University of Washington \\
        \And
      Mitchell Wortsman \\
    University of Washington \\
        \And
      Kiana Ehsani$^\dagger$ \\
    University of Washington\\
}

\begin{document}

\maketitle

\begin{abstract}
Although sparse neural networks have been studied extensively, the focus has been primarily on accuracy. In this work, we focus instead on \emph{network structure}, and analyze three popular algorithms. We first measure performance when structure persists and weights are reset to a different random initialization, thereby extending Zhou \textit{et al.} \cite{zhou2019deconstructing}. This experiment reveals that accuracy can be derived from structure alone. %Moreover, the structure can be analysed to reveal properties of the data. 
Second, to measure structural robustness we investigate the \emph{sensitivity} of sparse neural networks to further pruning after training, finding a stark contrast between algorithms. Finally, for a recent dynamic sparsity algorithm we investigate how early in training the structure emerges. We find that even after one epoch the structure is mostly determined, allowing us to propose a more efficient algorithm which does not require dense gradients throughout training.  
In looking back at algorithms for sparse neural networks and analyzing their performance from a different lens, we uncover several interesting properties and promising directions for future research.
\end{abstract}

\section{Introduction}
Sparse networks achieve impressive accuracy while using only a fraction of the parameters \cite{lecun1990optimal, han2015deep, gale2019state, frankle2018lottery, evci2019difficulty}. Though earlier work on sparse networks focused primarily on pruning \emph{after} training, researchers have recently shown interest in pruning early in training \cite{frankle2018lottery, lee2018snip, tanaka2020pruning, wang2019picking} or dynamically as training progresses \cite{dettmers2019sparse, wortsman2019discovering, evci2019rigging}.

Sparse neural networks continue to intrigue researchers for two main reasons. First, pruning neural networks can vastly reduce storage and computational costs, important steps towards reducing energy consumption and GreenAI \cite{schwartz2019green}. Additionally, sparsifying neural networks provides a useful tool for understanding learning dynamics and other phenomena. For instance, the \emph{lottery ticket hypothesis} \cite{frankle2018lottery} conjectures that overparameterized neural networks contain subnetworks which can be trained to competitive accuracy when their weights are reset to their initialization. These findings may suggest that SGD seeks out and trains a lucky subnetwork early in training. Morever, in \cite{zhou2019deconstructing,ramanujan2019s} high accuracy is achieved by sparsifying a randomly initialized neural network with fixed weights. 

In this paper we investigate sparse neural networks \emph{structures}. We are inspired by the analysis of \cite{zhou2019deconstructing}, wherein Zhou \textit{et al.} carefully analyze which aspects of the Iterative Magnitude Pruning (IMP) \cite{frankle2018lottery} contribute to the accuracy. We bring a similar tool and perspective to the analysis of network structure across different algorithms. In addition to IMP we analyze two dynamic sparsity methods: Discovering Neural Wirings (DNW) \cite{wortsman2019discovering} and Rigged Lottery Tickets (RigL) \cite{evci2019rigging}, algorithms where the connectivity changes throughout training. We then ask the following three questions:
\begin{enumerate}
    \item What role does \emph{structure} play in a sparse neural network's performance? As in \cite{zhou2019deconstructing}, we investigate this by using the sparse neural network structure found at the end of training and resetting the weights to a different random initialization before retraining. Our findings match \cite{zhou2019deconstructing} for IMP subnetworks, which perform similarly to random subnetworks when retrained with new randomly initialized parameters. Surprisingly, we find that this isn't true for DNW and RigL subnetworks, which outperform random subnetworks across various sparsity values.
    \item How robust are different network structures? Specifically, we investigate how quickly accuracy degrades when further pruning networks after training. We conclude that IMP, DNW, and RigL subnetworks are less sensitive than a random subnetwork when further pruning by magnitude.
    \item In DNW, a dynamic sparsity method, how early in training does the final structure emerge? We find that surprisingly the structure emerges fairly early in training and leverage these results to design a more efficient algorithm.
\end{enumerate}

%Network overparameterization is demonstrated in the Lottery Ticket hypothesis, which outlines a method of selecting trainable sparse subnetworks. The task of efficiently choosing a network with the fewest parameters while maintaining optimal accuracy is an open question in deep learning. One major complication is that different models with the same number of parameters can behave differently depending on the graph structure formed by their active parameters.

%Popular deep learning techniques such as convolutions and pooling take advantage of spatial relations to reduce the number of parameters necessary for learning. These techniques are implemented as network structures which are selected by hand using a process of informed trial and error. More recently, \textit{dynamic} techniques for training neural networks have opened an entire new realm of sparse network structures.

% more

\section{Preliminaries and Related Work}

We investigate four classes of sparse graphs. For the first class of graphs, which we refer to as a random graph, we choose a random set of weights to remain zero throughout training. We also consider Lottery Ticket (LT)~\cite{frankle2018lottery} graphs,  DNW~\cite{wortsman2019discovering} graphs, and RigL graphs~\cite{evci2019rigging} which are respectively produced by the following three algorithms:

\textbf{1. Iterative Magnitude Pruning (IMP).} IMP \cite{frankle2018lottery, frankle2020training} iteratively sparsifies a neural network in successive training runs. First a dense network is trained until completion, at which point the bottom $p\%$ of the weights by magnitude are set to 0. The nonzero weights are then reset to their initialization and the process continues until the requisite sparsity is attained. Although the process is expensive, IMP produces highly effective sparse neural networks.

\textbf{2. Discovering Neural Wirings (DNW).} As in \cite{dettmers2019sparse, evci2019rigging}, DNW \cite{wortsman2019discovering} maintains sparsity throughout training. In the forwards pass, DNW selects the top $p\%$ of weights by layer to be nonzero. In the backwards pass, all weights are updated via the straight-through estimator allowing the re-entry of dead weights.

\textbf{3. Rigged Lottery (RigL).} RigL \cite{evci2019rigging} maintains sparsity throughout training and does not require dense gradients during most iterations of training. RigL graph structure is static for the majority of training iterations, and only changes every $n$ iterations up to some stopping iteration $T_{end}$. On those iterations where graph structure \textit{is} changed, the bottom $k$ parameters are removed from the network by absolute magnitude. To replace these $k$ lost parameters, the $k$ parameters with the greatest magnitude gradient are introduced into the network. $k$ decreases as the number of iterations increases, until $T_{end}$ is reached.

\section{How Does Structure Affect Accuracy?}

The four algorithms we investigate find sparse neural graph structures using different methods. These differences between algorithms might be expected to lead each algorithm to generate graph structures with unique characteristics. Understanding how much graph \textit{structure} is responsible for the accuracy of each of these algorithms, as opposed to the importance of a specific initialization, is important for understanding the generalizability of the structures they create.

% The impact of graph structure on accuracy is broadly recognized within machine learning. Larger, deeper networks outperform smaller shallow networks in more complex classification tasks. It is less investigated how different sparse network structures perform when their weights are reset to random. This captures how much of the performance comes from the structure.

% The Lottery Ticket hypothesis presents a method for identifying subnetworks whose structure and parameter initialization allow for training even in sparse regimes. The Lottery Ticket hypothesis and works revolving around it have all relied to some degree on the initialization of parameters in the selected subnetwork. \todo{There was the paper with the +/- resetting of graph weights}\mitch{\cite{zhou2019deconstructing}}. 

To determine how the four different classes of graph structures perform under varying levels of sparsity, we compare their accuracy at each degree of sparsity when their parameters are reinitialized with multiple new random seeds. This experiment extends \cite{zhou2019deconstructing}, where authors conduct this analysis for LT graphs.

\begin{figure}[t!]
    \centering
    \includegraphics[width=\linewidth]{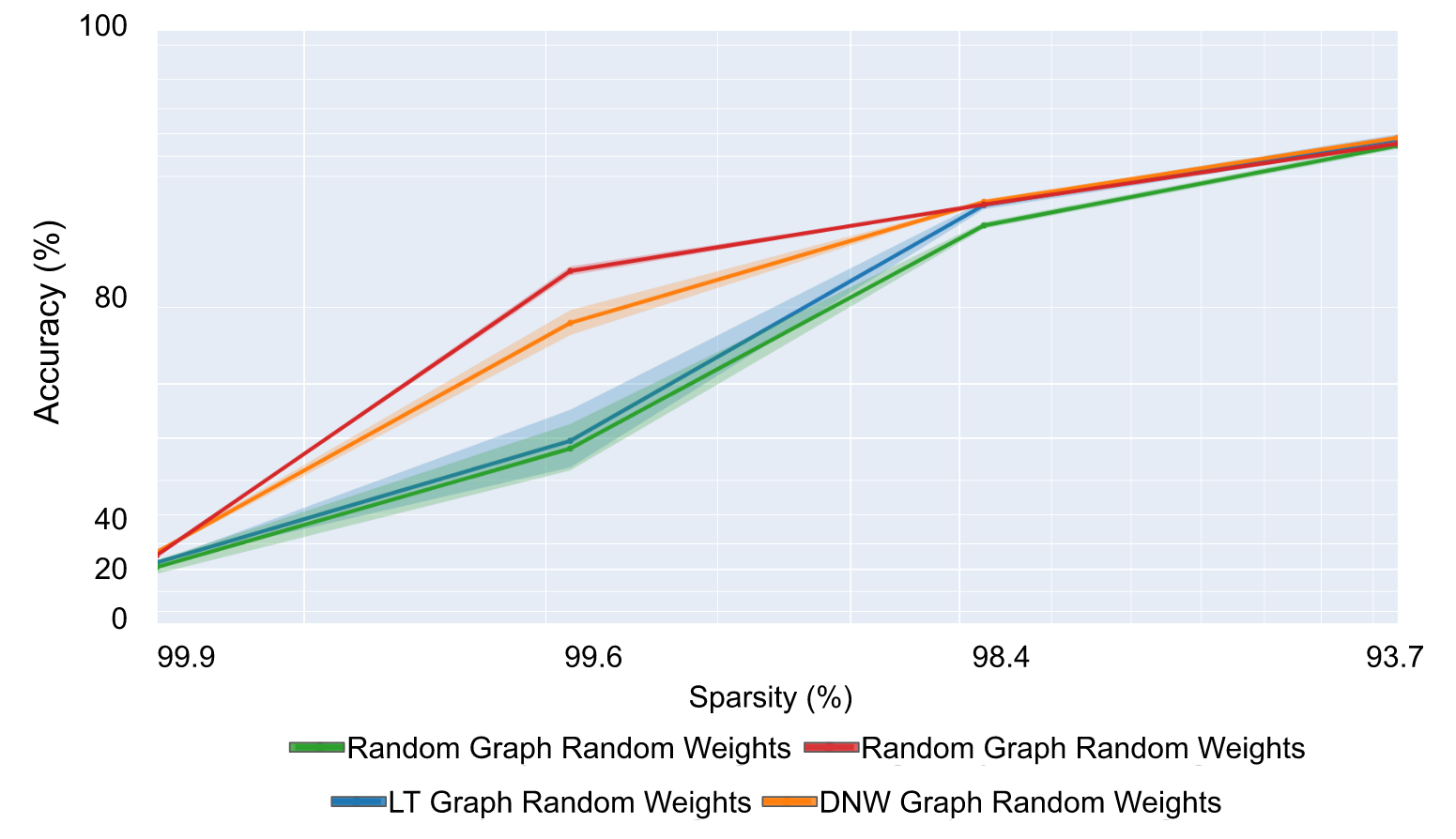}
    \caption{\textbf{Accuracy Vs. Sparsity.} Performance of ResNet-18 on CIFAR-10 when trained with the structure found by sparsity algorithms but with weights reset randomly. Error bars show standard deviation of accuracy when graph structures of each type are reinitialized with three random seeds.}
    \label{fig:sec3fig}
\end{figure}

Using ResNet-18 \cite{he2016deep} on CIFAR-10 \cite{cifar}, we find that LT graphs perform similarly to random graphs when their weights are reset randomly. This corroborates the findings of \cite{zhou2019deconstructing}. Surprisingly, however, we find that DNW and RigL graphs outperform random graphs, suggesting that their performance may in part due to a characteristic of their structure which is absent in LT and random graphs. As illustrated in Figure~\ref{fig:sec3fig}, this performance gap is especially pronounced in the sparse regimes.

\section{Sensitivity to Further Pruning}

In linear optimization, \textit{sensitivity} analysis can be used to identify the stability of a solution for maximizing or minimizing a linear program. When small perturbations to the inequalities defining a linear program leave the optimal solution unchanged, the optimal solution is said to be stable. In order to test the robustness of a graph's structure, we experiment with an analogous metric which measures the stability of a sparse neural network.

\begin{figure}[ht!]
\centering
\begin{subfigure}{\textwidth}
  \centering
  \includegraphics[width=\linewidth]{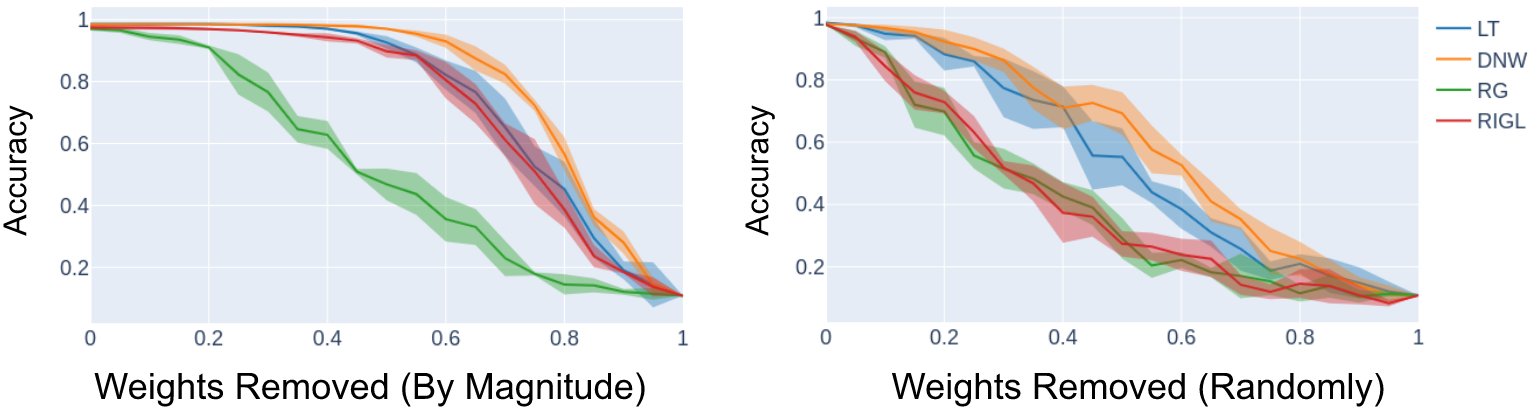}
\end{subfigure}
% \begin{subfigure}{.5\textwidth}
%   \centering
%   \includegraphics[width=\linewidth]{randomabblation.png}
% \end{subfigure}%
% \begin{subfigure}{.5\textwidth}
%   \centering
%   \includegraphics[width=\linewidth]{sensitivity.png}
% \end{subfigure}
\caption{Sensitivity to parameter zeroing for a 90.5\% sparse fully connected network on MNIST where weights are removed in order of ascending magnitude \textbf{(left)} and randomly \textbf{(right)}. Error bars show standard deviation with three random seeds.}
\label{fig:sense}
\end{figure}

To determine the robustness of each graph type, we introduce an algorithm for sensitivity analysis on neural graph structures. This sensitivity analysis is performed by further eliminating non-zero parameters \textit{after training} and evaluating the performance of the reduced network. We select which weights to remove both randomly and by order of ascending magnitude, as illustrated in Figure~\ref{fig:sense}. Our findings indicate that when removing weight by magnitude the performance of random graphs decreases faster than DNW, LT, and RigL graphs. This suggests that lower magnitude parameters in LT, DNW and Rigl graphs are less critical. This is partially expected for DNW and RigL, which continuously swap out the lower magnitude weights. When removing weights randomly, DNW and LT slightly outperform RigL and random graphs, suggesting that there may be more redundancy within LT and DNW graphs.

\section{How Early in Dynamic Training Does Structure Emerge?}

In order to improve dynamic training algorithms, it is helpful to have tools which monitor how they modify graph structure as they train. One simple metric which allows us to roughly track graph evolution during training is to track node in-degrees. By tracking the in-degree of each node in a layer we are able to gain insight into how networks evolve during training with dynamic algorithms. Surprisingly, when analysing DNW, we find that the majority of the structural changes happen in the first epoch, and that the remainder of the structural change is mainly an amplification of the structure created in the first epoch. Nodes which have higher in-degrees after the first epoch of training continue to increase the number of connections they have, while those nodes with lower in-degrees after the first epoch see declines in their importance over the rest of training.

\begin{figure}[ht!]
\centering
\begin{subfigure}{.5\textwidth}
  \centering
  \includegraphics[width=\linewidth]{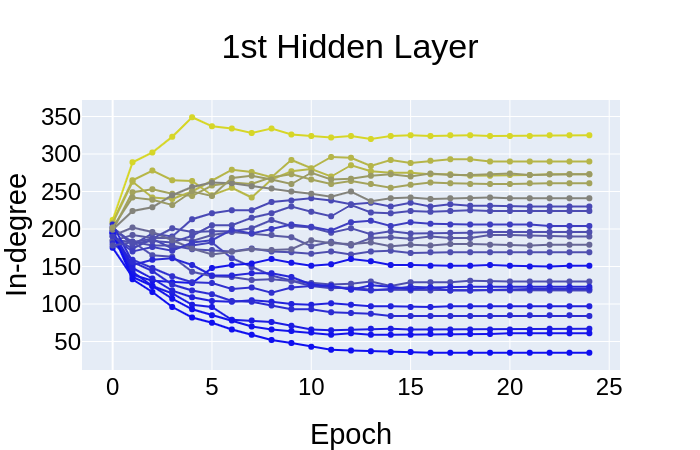}
\end{subfigure}%
\begin{subfigure}{.5\textwidth}
  \centering
  \includegraphics[width=\linewidth]{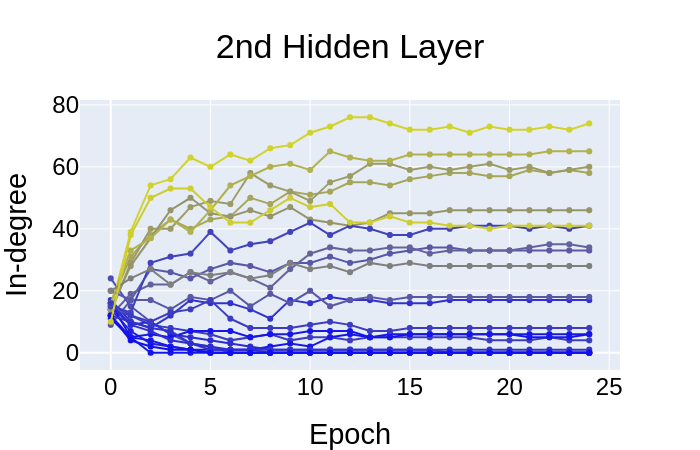}
\end{subfigure}
\caption{In-degree evolution throughout training with DNW. The in-degree of a random subsampling of neurons is shown for a small sparsified fully connected network on MNIST with 90.5\% sparsity. Color is used to show in-degree after one epoch of training, where yellow is a higher in-degree after the first epoch and blue is a lower in-degree after the first epoch. These results showcase a strong correlation between in-degree after the first epoch and final in-degree.}
\label{fig:degree-evolution}
\end{figure}

For normal DNW training, the gradient is propagated to all parameters with the straight through estimator \cite{bengio2013estimating}, allowing dead weights to resurface. In a very sparse graph, this means that a high percentage of the computational power is consumed looking for new edges to add to the graph. However, we have shown in Figure~\ref{fig:degree-evolution} that the final DNW structure is primarily an amplification of the structure after the first epoch, and therefore much of this searching is superfluous.

\begin{figure}[ht!]
\centering
\begin{subfigure}{\textwidth}
    \centering
    \includegraphics[width=\linewidth]{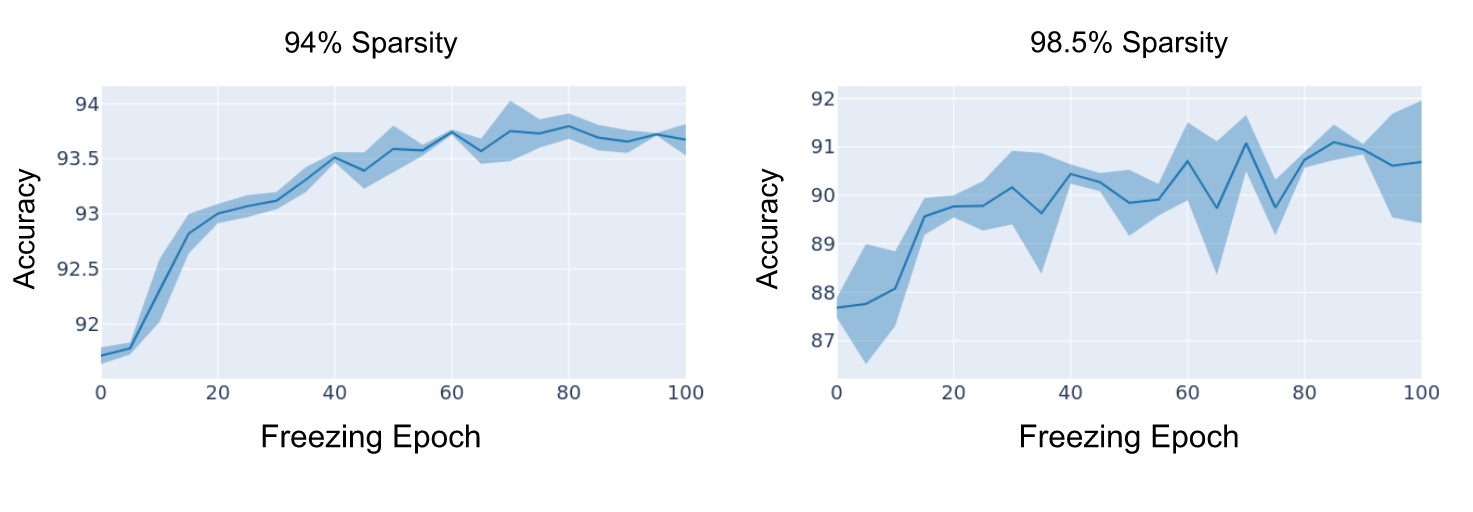}
\end{subfigure}
% \begin{subfigure}{.5\textwidth}
%   \centering
%   \includegraphics[width=\linewidth]{freeze025.png}
% \end{subfigure}%
% \begin{subfigure}{.5\textwidth}
%   \centering
%   \includegraphics[width=\linewidth]{freeze0125.png}
% \end{subfigure}
%\vspace{-1em}
\caption{The effect of the \emph{freezing} epoch on performance in our hybrid dynamic/static training method for ResNet-18 on CIFAR-10. Error bars give standard deviation from three random seeds at each freezing epoch.}
\label{fig:hybrid}
\end{figure}

Based on these results, we propose a hybrid dynamic/static training method where the structure is allowed to change in the first $k$ epochs using the normal DNW algorithm, and structure is fixed for the remainder of training. Accordingly we refer to the $k$th epoch by the \emph{freezing} epoch. We illustrate the performance of our hybrid algorithm in Figure~\ref{fig:hybrid}.

We find that, after the first few epochs, further training with DNW yields diminishing returns. This corroborates our finding that structure is primarily determined within the first few epochs of DNW. However, if compute is unbounded it is still best to train with DNW to completion. In comparison to newer dynamic training algorithms including RigL and Top-KAST \cite{evci2019rigging, jayakumar2020top}, our modified DNW algorithm is likely still less computationally efficient for most applications. We believe the more important finding, however, is that simple introspection into graph evolution during dynamic training can make improvements to existing algorithms easier to identify.

\section{Conclusion} \label{sec:conclusion}

We have taken a retrospective look at sparse neural networks through a different lens. Our findings offer interesting insight into a growing field which is becoming increasingly practical \cite{gale2020sparse}. We have shown that in contrast with IMP, structures found by DNW and RigL perform well independently of initialization. This suggests that there are properties in dynamically generated structures that are absent in LT and random graphs which are conducive to sparse learning. Moreover we have shown that random graphs are much more sensitive to changes in connectivity than the other graph types. Finally, using a simple tool we find that the majority of structural changes in DNW occur in the first few epochs, leading us towards a more efficient algorithm. More generally, we believe that analyzing existing algorithms from a variety perspectives is often very useful and underexplored.

%\mitch{TODO read conclusion}
%Sparse neural networks find proponents in those who seek to increase computational efficiency, mimic biological brains, and many others \maxvg{weak!}. Our findings show that graphs in the classes \textbf{DNWG} and \textbf{LTG} outperform graphs in the class \textbf{RG} in sparse regimes independently of initialization. Moreover, we find that \textbf{RG} graphs are more sensitive to small quantities of pruning than graphs in \textbf{DNWG} and \textbf{LTG}. 

{
\bibliographystyle{plain}
\bibliography{main.bib}

\begin{thebibliography}{10}

\bibitem{bengio2013estimating}
Yoshua Bengio, Nicholas L{\'e}onard, and Aaron Courville.
\newblock Estimating or propagating gradients through stochastic neurons for
  conditional computation.
\newblock {\em arXiv preprint arXiv:1308.3432}, 2013.

\bibitem{dettmers2019sparse}
Tim Dettmers and Luke Zettlemoyer.
\newblock Sparse networks from scratch: Faster training without losing
  performance.
\newblock {\em arXiv preprint arXiv:1907.04840}, 2019.

\bibitem{evci2019rigging}
Utku Evci, Trevor Gale, Jacob Menick, Pablo~Samuel Castro, and Erich Elsen.
\newblock Rigging the lottery: Making all tickets winners.
\newblock {\em arXiv preprint arXiv:1911.11134}, 2019.

\bibitem{evci2019difficulty}
Utku Evci, Fabian Pedregosa, Aidan Gomez, and Erich Elsen.
\newblock The difficulty of training sparse neural networks.
\newblock {\em arXiv preprint arXiv:1906.10732}, 2019.

\bibitem{frankle2018lottery}
Jonathan "Frankle and Michael Carbin.
\newblock "the lottery ticket hypothesis: Finding sparse, trainable neural
  networks.
\newblock {\em "arXiv preprint arXiv:1803.03635}, "2018.

\bibitem{frankle2020training}
Jonathan "Frankle, David~J Schwab, and Ari~S Morcos.
\newblock "training batchnorm and only batchnorm: On the expressive power of
  random features in cnns.
\newblock {\em "arXiv preprint arXiv:2003.00152}, "2020.

\bibitem{gale2019state}
Trevor Gale, Erich Elsen, and Sara Hooker.
\newblock The state of sparsity in deep neural networks.
\newblock {\em arXiv preprint arXiv:1902.09574}, 2019.

\bibitem{gale2020sparse}
Trevor Gale, Matei Zaharia, Cliff Young, and Erich Elsen.
\newblock Sparse gpu kernels for deep learning.
\newblock {\em arXiv preprint arXiv:2006.10901}, 2020.

\bibitem{han2015deep}
Song Han, Huizi Mao, and William~J Dally.
\newblock Deep compression: Compressing deep neural networks with pruning,
  trained quantization and huffman coding.
\newblock {\em arXiv preprint arXiv:1510.00149}, 2015.

\bibitem{he2016deep}
Kaiming He, Xiangyu Zhang, Shaoqing Ren, and Jian Sun.
\newblock Deep residual learning for image recognition.
\newblock In {\em Proceedings of the IEEE conference on computer vision and
  pattern recognition}, pages 770--778, 2016.

\bibitem{jayakumar2020top}
Siddhant Jayakumar, Razvan Pascanu, Jack Rae, Simon Osindero, and Erich Elsen.
\newblock Top-kast: Top-k always sparse training.
\newblock {\em Advances in Neural Information Processing Systems}, 33, 2020.

\bibitem{cifar}
"Alex Krizhevsky.
\newblock "learning multiple layers of features from tiny images.
\newblock Technical report, "University of Toronto, "2009.

\bibitem{lecun1990optimal}
Yann LeCun, John~S Denker, and Sara~A Solla.
\newblock Optimal brain damage.
\newblock In {\em Advances in neural information processing systems}, pages
  598--605, 1990.

\bibitem{lee2018snip}
Namhoon Lee, Thalaiyasingam Ajanthan, and Philip Torr.
\newblock Snip: Single-shot network pruning based on connection sensitivity.
\newblock In {\em International Conference on Learning Representations}, 2018.

\bibitem{ramanujan2019s}
Vivek "Ramanujan, Mitchell Wortsman, Aniruddha Kembhavi, Ali Farhadi, and
  Mohammad Rastegari.
\newblock "what's hidden in a randomly weighted neural network?
\newblock {\em "arXiv preprint arXiv:1911.13299}, "2019.

\bibitem{schwartz2019green}
Roy Schwartz, Jesse Dodge, Noah~A Smith, and Oren Etzioni.
\newblock Green ai. corr abs/1907.10597 (2019).
\newblock {\em arXiv preprint arXiv:1907.10597}, 2019.

\bibitem{tanaka2020pruning}
Hidenori Tanaka, Daniel Kunin, Daniel~LK Yamins, and Surya Ganguli.
\newblock Pruning neural networks without any data by iteratively conserving
  synaptic flow.
\newblock {\em arXiv preprint arXiv:2006.05467}, 2020.

\bibitem{wang2019picking}
Chaoqi Wang, Guodong Zhang, and Roger Grosse.
\newblock Picking winning tickets before training by preserving gradient flow.
\newblock In {\em International Conference on Learning Representations}, 2019.

\bibitem{wortsman2019discovering}
Mitchell Wortsman, Ali Farhadi, and Mohammad Rastegari.
\newblock Discovering neural wirings.
\newblock In {\em Advances in Neural Information Processing Systems}, pages
  2684--2694, 2019.

\bibitem{zhou2019deconstructing}
Hattie "Zhou, Janice Lan, Rosanne Liu, and Jason Yosinski.
\newblock "deconstructing lottery tickets: Zeros, signs, and the supermask.
\newblock In {\em "Advances in Neural Information Processing Systems}, pages
  "3592--3602, "2019.

\end{thebibliography}
}

% \appendix

% \section{Additional Figures}

\end{document}